\documentclass[conference]{IEEEtran}
\IEEEoverridecommandlockouts
\usepackage{cite}
\usepackage{amsmath,amssymb,amsfonts}
\usepackage{algorithmic}
\usepackage{graphicx}
\usepackage{textcomp}
\usepackage{xcolor}
\usepackage{authblk}
\usepackage{hyperref}
\usepackage{algorithm} 
\usepackage{multirow}
\usepackage{array}
\usepackage{booktabs}
\usepackage{array}
\newcolumntype{C}[1]{>{\centering\let\newline\\\arraybackslash\hspace{0pt}}m{#1}}
\usepackage{dblfloatfix}

\newcommand{\etal}{\textit{et al.}}

\def\BibTeX{{\rm B\kern-.05em{\sc i\kern-.025em b}\kern-.08em
    T\kern-.1667em\lower.7ex\hbox{E}\kern-.125emX}}
\begin{document}

\title{Household navigation and manipulation for everyday object rearrangement tasks
}
%Household navigation ....

\author[2]{Shrutheesh R. Iyer*\thanks{*Work done while at the Contextual Robotics Institute, UC San Diego}} 
\author[1]{Anwesan Pal}
\author[1]{Jiaming Hu}
\author[1]{Akanimoh Adeleye}
\author[1]{Aditya Aggarwal}
\author[1]{Henrik I. Christensen}
\affil[1]{Contextual Robotics Institute, UC San Diego}
\affil[2]{Aurora Operations, Inc.}
\affil[1,2]{\texttt{\{siyer, a2pal, jih189, akadeley, a9aggarwal, hichristensen\}@ucsd.edu}}

\maketitle

\begin{abstract}
We consider the problem of building an assistive robotic system that can help humans in daily household cleanup tasks. Creating such an autonomous system in real-world environments is inherently quite challenging, as a general solution may not suit the preferences of a particular customer. Moreover, such a system consists of multi-objective tasks comprising -- (i) Detection of misplaced objects and prediction of their potentially correct placements, (ii) Fine-grained manipulation for stable object grasping, and (iii) Room-to-room navigation for transferring objects in unseen environments. This work systematically tackles each component and integrates them into a complete object rearrangement pipeline. To validate our proposed system, we conduct multiple experiments on a real robotic platform involving multi-room object transfer, user preference-based placement, and complex pick-and-place tasks. Additional details including video demonstrations of our work are available at \url{https://sites.google.com/eng.ucsd.edu/home-robot}.
\end{abstract}

\begin{IEEEkeywords}
long-term navigation, real-world manipulation, preferential object placement
\end{IEEEkeywords}

\section{Introduction}
Creating autonomous agents to aid human beings in everyday household chores has long been considered to be the holy grail of service robotics research. In this work, we take a step towards that goal by proposing a complete system for an indoor tidy-up task. Usually, this comprises of identifying misplaced objects in the environment, and transferring them to their desired locations. Several aspects of this inherently long-horizon task make it particularly challenging in a real-world environment. Firstly, recognizing out of place objects in a noisy environment is a non-trivial problem. While state-of-the-art open-vocabulary object detectors \cite{zareian2021open, gu2021open, minderer2022simple, zhou2022detecting} are quite adept at localizing objects in a zero-shot manner, determining whether they belong in a particular environment is more complicated, as it also involves understanding scene context. Secondly, user preferences for placing objects in the ``correct" room and surface (hereafter called \textit{receptacle}), are often subjective, thereby inhibiting the sole use of generic common-sense reasoning models. Thirdly, manipulating unknown objects in a cluttered environment is still an open research problem due to the difficulty of affordance estimation and motion planning. Finally, delivering an object to a previously unlabeled receptacle in the target room is particularly challenging, specially if the precise location of said receptacle is unknown.

% to include collaborative filtering.
% contributions are : filtering for common sense knowledge and human preferences in a knowledge graph structure.
% delivering to unannotated target receptacles
% complex manipulation task

\begin{figure}[t]
    \centering
    \includegraphics[width=0.45\textwidth]{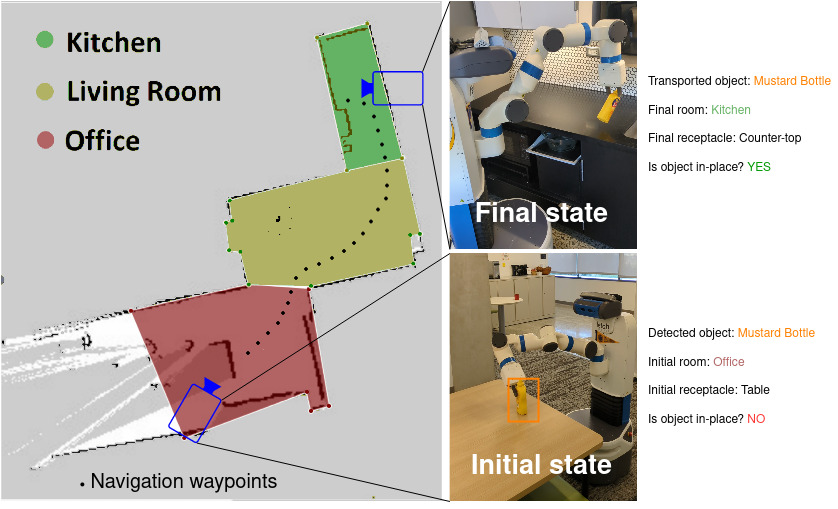}
    \caption{
    An example of a home-robot rearrangement task. At the initial state, the robot identifies the $\mathtt{mustard\_bottle}$ object and determines that it is misplaced in the $\mathtt{office}$. Subsequently, the robot transports it to its correct location in the $\mathtt{kitchen}$ on top of the $\mathtt{counter\text{-}top}$. The semantic map used for the navigation task is shown on the left with the robot's trajectory.
    % An example of home-robot rearrangement task. Initially, the robot detects an $\mathtt{mustard\_bottle}$ and determines that it is misplaced on the $\mathtt{office}$ $\mathtt{table}$. A reasoning model (shown on the right side), then predicts that $\mathtt{mustard\_bottle}$ should be placed on the $\mathtt{counter\text{-}top}$ of a $\mathtt{kitchen}$. Next, the robot picks it up and plans a path to the $\mathtt{counter\text{-}top}$ with the given constructed semantic map of the environment shown in the Left. Upon reaching the $\mathtt{kitchen}$, the robot will search for a possible $\mathtt{counter\text{-}top}$. Once localizing $\mathtt{counter\text{-}top}$, the robot will approach and place the $\mathtt{mustard\_bottle}$ onto it.
    }
    \label{fig:intro}
\end{figure} 

% In this paper, we meticulously address each component outlined above, seamlessly integrating them to form a comprehensive pipeline of the in-home rearrangement system. This pipeline facilitates the reorganization of household items in authentic real-world settings using the Fetch~\cite{fetch2016} mobile manipulation platform. Figure \ref{fig:intro} illustrates an example of a home-robot rearrangement task that underscores the scope of our study.

In this paper, we address each component for rearranging household objects in a real-world setting utilizing the Fetch~\cite{fetch2016} mobile manipulation platform. To ensure robustness and scalability within the physical world, we propose a modular system that is capable of performing (i) user-preference based reasoning through collaborative filtering, (ii) fine-grained pick-up of unknown objects and placement on previously unlabeled receptacles, and (iii) multi-room rearrangement. All these functionalities are coordinated by behavior trees that can handle failure at different levels. An example of the operation is shown in Figure~\ref{fig:intro}.

% As shown in Figure~\ref{fig:intro}, the robot detects a misplaced object ($\mathtt{mustard\_bottle}$) and the reasoning module predicts its correct placement ($\mathtt{counter\text{-}top}$ on $\mathtt{kitchen}$). Then, the misplaced object is placed on the target receptacle by navigating through the environment.

% At the initial state, the robot detects an object $\mathtt{mustard\_bottle}$, and determines that it is misplaced in the $\mathtt{office}$ environment atop a $\mathtt{table}$. Using a reasoning module, it further predicts a probable correct room and receptacle surface location for it. In this case, the predicted room is $\mathtt{kitchen}$, and the receptacle is $\mathtt{counter\text{-}top}$. Then, the robot picks up the misplaced object by stable grasping, and plans a path to the target room using a constructed semantic map of the environment. Upon reaching the $\mathtt{kitchen}$, the robot scans the surroundings for a possible $\mathtt{counter\text{-}top}$. Once it has succeeded in correctly localizing it, the robot plans a path to reach the $\mathtt{counter\text{-}top}$, and places the held $\mathtt{mustard\_bottle}$ safely on the receptacle.

The rest of the paper is organized as follows. Section \ref{sec: rel_work} discusses existing approaches for object rearrangement in home-robot environments. Section \ref{sec: components} has a description of each component we use to perform the overall task, with a summary of the integrated system in Section \ref{sec: sys_int}. We explain the conducted experiments in detail in Section \ref{sec: expt}, and provide a summary of our work with some future goals in Section \ref{sec: conclusion}.

\section{Related Work} \label{sec: rel_work} % Sanmi & Anwesan 
Recently, indoor object rearrangement tasks using mobile robots have received a lot of attention from the robotics and computer vision community. Due to the increasing number of Embodied AI platforms available \cite{puig2018virtualhome, shridhar2020alfred, shridhar2020alfworld, szot2021habitat, Berges_2023_CVPR, li2021igibson}, several approaches have been proposed for solving the complete mobile manipulation task in a number of home environments. However, most of these methods \cite{huang2022language, min2021film, puig2018virtualhome, batra2020rearrangement, szot2021habitat} are entirely trained in simulation, and therefore rarely generalize to real-world environments. Other works have adopted the task planning approach, but are either restricted to specific tasks such as folding clothes \cite{srivastava2015tractability} and rearranging kitchens \cite{wu2023m}, or follow a pre-defined template \cite{cui2021semantic}. Some approaches \cite{schaeffer1999care, graf2004care, reiser2009care, kittmann2015let} focus on the human-robot interaction aspect, but not on autonomy. Lately, large language models (LLMs) have gained popularity for robotic manipulation, both for task planning \cite{wu2023tidybot, ding2023task, chang2023lgmcts}, as well as end-to-end execution \cite{brohan2023rt1, jiang2023vima, brohan2023rt2, ahn2022i}. While these large foundational models are proficient at reasoning about object semantics, accurately grounding the offline acquired knowledge in a dynamic physical environment is still considered to be a non-trivial problem. Two efforts closest to ours are that of Wu \etal \cite{wu2023tidybot} and Castro \etal \cite{castro2022bt}. Wu \etal \cite{wu2023tidybot} use LLMs to infer generalized user preferences and use it to tidy a room. However, they do not handle fine-grained manipulation, need rigorous prompt engineering to understand user preferences, and are limited to within-room navigation. Castro \etal \cite{castro2022bt} do consider room-to-room navigation, but they rely on manually annotated prior semantic maps for querying the exact locations of target rooms and receptacles. In contrast, we only build a simple $2$D geometric map with rough room locations, and proceed to identify receptacles in the environment on the fly. 
% \subsection{Use of LLMs for robotics}
% \subsection{TidyBot}

\section{Components} \label{sec: components}

% In this section, we will go over the individual components that together make up our ensemble system for rearranging objects in a home environment.

Our proposed ensemble system for home-robot rearrangement contains four primary modules: scene recognition and mapping, object rearrangement, manipulation, and navigation. 

\subsection{Semantic mapping and visual recognition} \label{sec: perception}
Our detection module perceives the environment in two stages. The first stage involves construction of a semantic map of the environment for localization, while the second stage deals with recognition of objects in the environment. The localization system uses Cartographer mapper \cite{hess2016real} to generate a LiDAR-based $2$D occupancy-grid environment map. For simplicity, we manually annotate the locations in the map with a semantic label of the room category. This manual annotation step can also be replaced by an automated module such as \cite{pal2019deduce}. We do not annotate locations of receptacles however, as knowing their exact positions apriori is a strong assumption in dynamic environments. For object recognition, we use the DETIC~\cite{zhou2022detecting} model trained on twenty-thousand object classes. With this detector, we can detect both manipulable objects and receptacle surfaces for the rearrangement task.

% We use the detector in two-stages -- (i) To detect target objects that we can manipulate, where we extract $2$D bounding-boxes from the images for precise localization, and (ii) For detecting receptacle surfaces which are often irregularly shaped, where we extract segmentation masks which roughly identifies the region corresponding to receptacles.
% \begin{figure}[htp]
%     \centering
%     \includegraphics[width=0.5\textwidth]{images/mapping.png}
%     \caption{2D Mapping Process}
%     \label{fig:mapping}
% \end{figure}

\subsection{Object rearrangement} \label{sec:oop}

The rearrangement module involves repositioning objects in the home, using both \textit{common-sense reasoning} (to determine target rooms) and \textit{human preferences} (for selecting target receptacles).
% In this work, our motivation for rearranging homes is to do so based on both \textit{common-sense reasoning} (for target \textit{rooms}), and \textit{human preference} (for target \textit{receptacles}). 
We utilize a large human-labeled dataset~\cite{kant2022housekeep} for object placement preferences in homes, creating a knowledge base to predict likely room locations for objects. Then, we utilize user preference to capture diversity in human choices for receptacle locations.
% We leverage a recently proposed large dataset \cite{kant2022housekeep} of human-labeled preferences for object placement location in a home environment as our object vocabulary. Using this, we create a knowledge base of most likely room locations for a particular object based on a majority vote. This corresponds to common-sense reasoning of which rooms are likely candidates for hosting a particular object. For predicting target receptacle locations, we utilize user preference to learn diversity in human choices. 
However, the dataset does not contain a particular user's preference for \textit{all} the objects, leading to a sparse user-preference matrix. Thus, given scores and user ranked preferences, we use collaborative filtering~\cite{2015addo} to fill out the sparse matrix. Subsequently, matrix factorization \cite{koren2008factorization} is performed to predict user ratings ${r_{u,i}}$ for user $u$ and item $i$.
% We build on existing work using user preferences through collaborative filtering to put away groceries and toys~\cite{2015addo}. Collaborative filtering is a recommender system where, given scores for users ranked preferences, we can construct a sparse user matrix and perform matrix factorization \cite{koren2008factorization}. This allows us to predict user ratings ${r_{u,i}}$ for user $u$ and item $i$. 
In our case, an item $i$ refers to an object's placement in a particular room and receptacle. We can predict a user's rating using $f(u,i) = \gamma{_u} * \gamma{_i}$. Here, $\gamma_u \in \mathbb{R}^d$ and $\gamma_i \in \mathbb{R}^d$ are latent vectors representing the row of a user in matrix $\gamma_U$ and column of an item in matrix $\gamma_I$, and $d$ is the lower dimensional space.  To choose parameters $\gamma = \{ \gamma_u ,\gamma_i\}$ to closely fit the data, we minimize a loss function using Mean Squared Error with an L2 regularization term. 

\begin{equation}
\arg\min_{\gamma} \frac{1}{|\tau|} \sum_{r_{u,i} \in \tau}  w_{u,i}(r_{u,i} - f(u,i))^2  + \lambda\Omega(\gamma)
\end{equation}

where $\tau$ is our corpus of ratings and $\Omega(\gamma)$ is $\ell$2 norm $||\gamma||{_2^2}$. Our approach allows us to estimate the full preferences of users' desired correct object placement locations. 

Object rearrangement involves two main steps: (1) Identifying misplaced objects by checking if their current location is in the top-k (10 in this work) likely locations from our user-preference matrix, and (2) Predicting preference-based placement by first determining the target room using common-sense reasoning, and then identifying various potential receptacle locations within that room based on a sampled user identity.
% The overall task of object rearrangement involves primarily two steps: (i) misplaced object identification -- For this, we enumerate a list of top-$k$ ($k=10$ in this work) likely locations (room and receptacle) based on our completed user matrix. If the current room and receptacle location for a particular object does not fall in this list, we classify it as misplaced. (ii) Preference-based object placement prediction -- For every misplaced object, we first use common-sense reasoning to determine its target room location. Subsequently, the various preference-based target receptacle locations corresponding to that room can be obtained based on a sampled user identity.

% We combine this method with a general Knowledge Graph to filter out possible erroneous scores given by our matrix factorization. Our KG is comprised of .....
% Given this, if our user's top score for an object placement is not within the top 10 ranked room placements for our knowledge graph, the next highest score is used. This process repeats until we have an object placement in the top-scoring receptacle of our collaborative filtering method, such that the room is within the top-ranked placements of our KG for that object.  

%We change the traditional MSE loss function with regularizes to include an additional term that incorporates the cosine similarity of \textit{text-embedding-ada-00}'s \cite{} word embedding for an object given a room.

\begin{figure*}[t]
    \centering
    \includegraphics[width=0.9\textwidth]{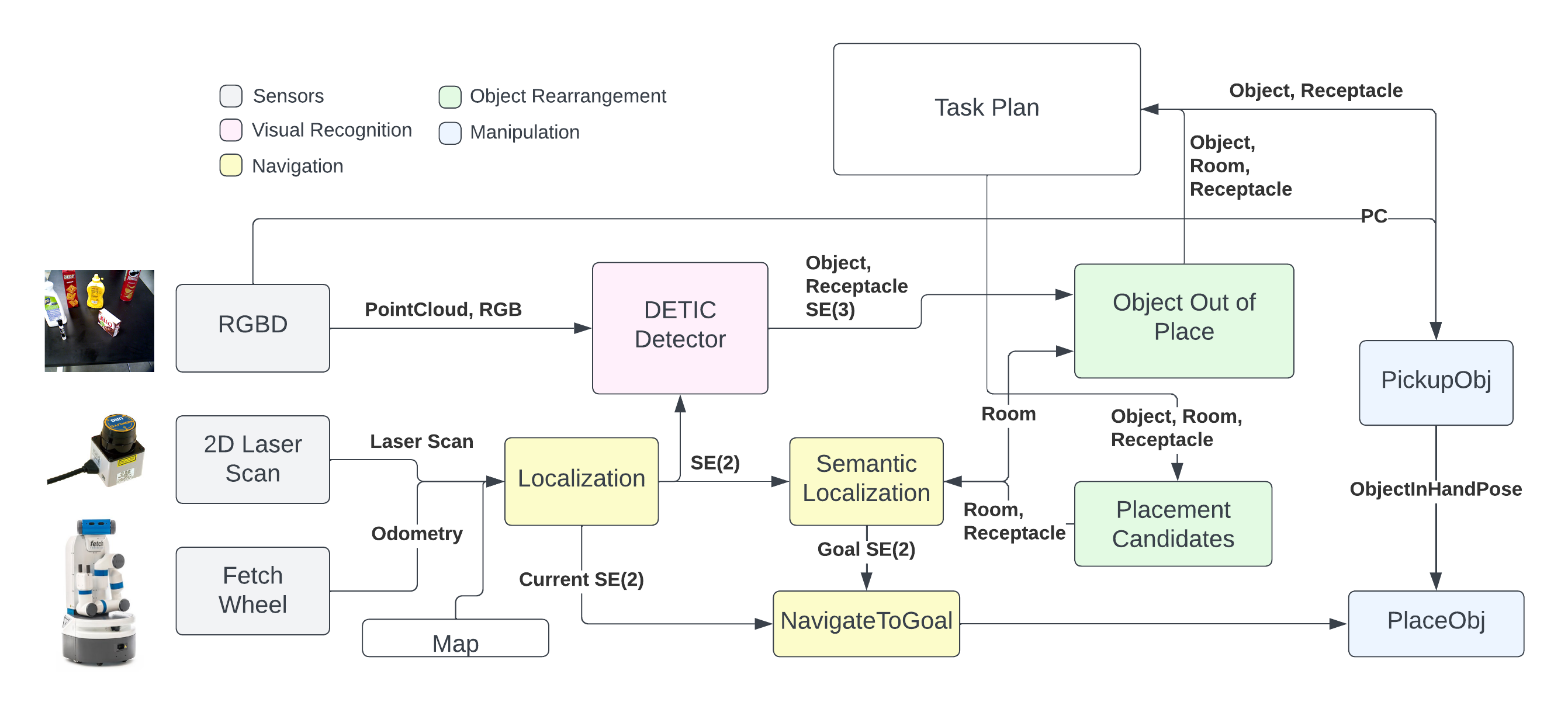}
    \caption{The overall architecture of the proposed system, as discussed in \ref{sys_arch} }
    \label{fig:overall_arch}
\end{figure*} 

\subsection{Manipulation of objects} \label{sec:manipulation}
% The manipulation module comprises of three essential components. Firstly, planning involves understanding the environment and constructing a comprehensive planning scene. Secondly, it requires to carefully analyzing potential methods of interaction with the target object. Lastly, it needs to meticulously plan the motion required to effectively interact with the object, all while keeping the task goal in mind.

The manipulation module includes planning to understand and construct a scene, analyzing interaction methods with the target object, and planning the required motion for effective interaction, all aligned with the task goal.

Before constructing the planning scene, the robot in this work possesses some prior knowledge of the environment. For instance, it understands that most objects should be positioned on a flat receptacle such as a $\mathtt{table}$, or $\mathtt{counter\text{-}top}$. Therefore, the receptacle serves as a common obstacle during our manipulation tasks, making it beneficial to prioritize its search once an object is detected. Finally, the receptacle is added as a single entity in the planning scene for efficient collision detection, while a voxel set represents the remaining non-target objects, optimizing resource usage.

Even though the robot knows the planning scene, interacting with the target object is crucial. In this work, grasping is the prevailing contact approach. For this, a learning-based grasp prediction \cite{sundermeyer2021contact} model is utilized to estimate a set of possible grasping poses. However, pick-and-place is not the only manipulation action available. The robot must also account for potential object motions based on the task requirements. For instance, it might need to open a drawer before placing an object. Consequently, the robot must compute the required motion to open it after identifying a set of arm configurations to grasp the drawer handle. The robot may explore alternative approaches if the motion is found before the timeout.

% During actual motion-planning, the robot must consider various constraints that may arise, such as delivering a cup of water or opening a drawer. Hence, we employ CBIRRT~\cite{Berenson2009cbirrt} to address these constraints effectively.

\subsection{Semantic navigation} \label{sec: nav}
% The overall movement of the robot in our home environment is considered in two stages 
% The navigation module aims to move the robot between different locations for the rearrangement task, and is considered in two stages -- (i) room-to-room navigation, for planning the path from a source room to the target room, and (ii) receptacle navigation, for navigating to the correct receptacle in the target room for object placement. 
The navigation module aims to move the robot between different locations for the rearrangement task, and is considered in two stages -- (i) room-to-room navigation for planning a path to the target room, and (ii) receptacle navigation for navigating to the correct receptacle in the target room. 

For room-to-room navigation, the $2$D coordinate of the center of the target room is first computed from the annotated semantic map. Using this destination point, a heuristic point-goal navigation algorithm is adopted to plan a trajectory by avoiding obstacles along the way with the Navfn planner. Upon reaching the target room, the receptacle navigation module is called. First, the entire room is scanned for possible receptacles for the held object, and the position of each candidate receptacle is updated in the map by re-projecting the detected object from the depth map of the camera. Then, the most likely target receptacle is chosen as per the rearrangement module \ref{sec:oop}. Finally, a second heuristic planner is called to make the robot move as close to the goal receptacle position as is feasible in collision-free space, which is achieved through the Carrot Planner. %todo(siyer) : cite these

\section{System Integration} \label{sec: sys_int}

% Integrating all the aforementioned modules is a non-trivial task. To address this, we first outline the primary structure of the system. Subsequently, we will elucidate how behavior trees assume a pivotal role in deploying each module, both in detecting misplaced objects and in successfully executing the rearrangement task. Eventually, we will briefly explain how our system interfaces with the robot using ROS.
In this section, we first outline the primary structure of our proposed system, and then discuss the flow of control using behavior trees.

\subsection{System Architecture} \label{sys_arch}% Task Setup and Flow

Figure \ref{fig:overall_arch} depicts the overall architecture of the proposed system. The task plan is provided in the form of behavior trees, as discussed in the next section. The localization module reads the semantic map, along with sensor data, to get the robot's current coordinates in the room. The detector module reads the sensor data, along with the robot's location, and identifies objects in the environment along with their 3D locations in the map. The object rearrangement module obtains a list of $(object, receptacle, room)$ tuples from the perception and localization modules to identify misplaced objects and propose ``correct" placements. The manipulation module picks up the misplaced object. The target room for placement provides the goal location for the navigator module, which then calls the perception module to locate the target receptacle and navigate to it. The manipulation module finally places the object either on the receptacle or inside the receptacle, depending on the specified goal from the rearrangement module.

% The entire system is written within the ROS framework. Each module is written as an independent ROS node, and exposed either as a ROS service or a ROS action node.

% Task Setup
\subsection{Use of Behavior Trees for Integration}
A key component of the complex home-robot system is the composition of the different capabilities of the robot to execute the task robustly and continuously. This calls for a control architecture that is modular and capable of switching between tasks such that the different tasks can be called anywhere during the workflow. Consequently, Behavior Trees (BTs) are used to monitor and orchestrate the flow of the entire system. 
BTs is a modular control architecture developed for controlling autonomous agents that supports reactive behavior. \cite{colledanchise2018behavior} A BT consists of control nodes and leaf nodes, where the leaf nodes are atomic operations that include actuation and sensing. The control nodes are behavior nodes that chain together multiple nodes. Each node (with its children) is a behavior that the robot can exhibit. A behavior can be composed of multiple behaviors. For instance, picking up a misplaced object is composed of two behaviors: identifying a misplaced object and picking up a target object. 
\begin{figure*}[!b]
    \renewcommand{\thefigure}{4}
    \centering
    \includegraphics[width=\textwidth]{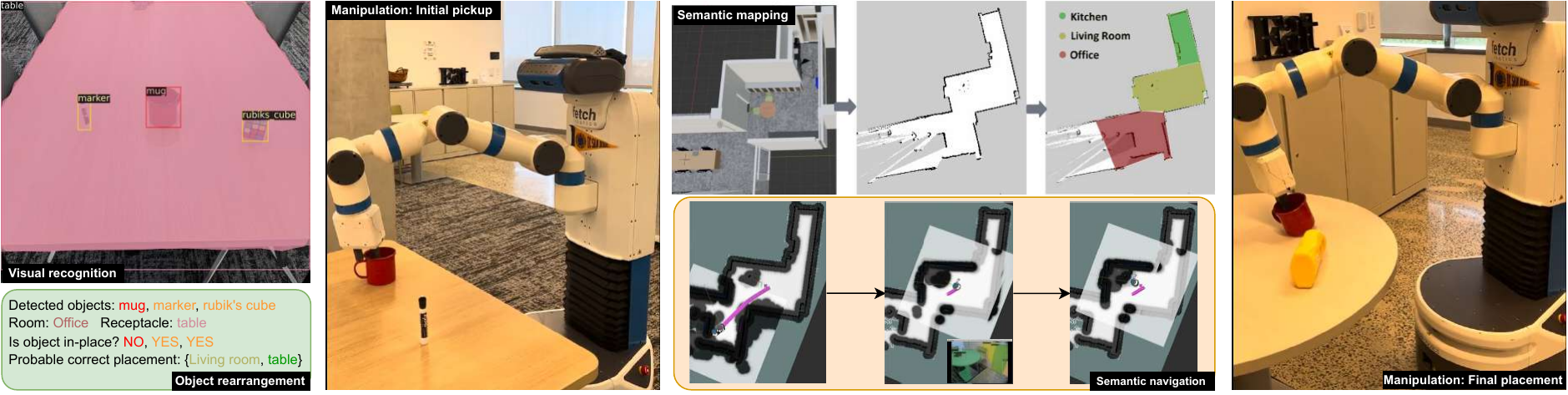}
    \vspace{-5mm}
    \caption{All proposed system components. Visual recognition module detects both target objects and receptacle surfaces. Object rearrangement module identifies misplaced objects and suggest their desired location. The manipulation module ensures the reliability of each pick and place action. Mapping module builds a $2$D environment map and semantically paints it with room labels. Finally, the navigation module uses the semantic map to plan the robot's trajectory.}
    \label{fig:all_comps}
\end{figure*} 

\begin{figure}[t]
    \renewcommand{\thefigure}{3}
    \centering
    \includegraphics[width=0.35\textwidth]{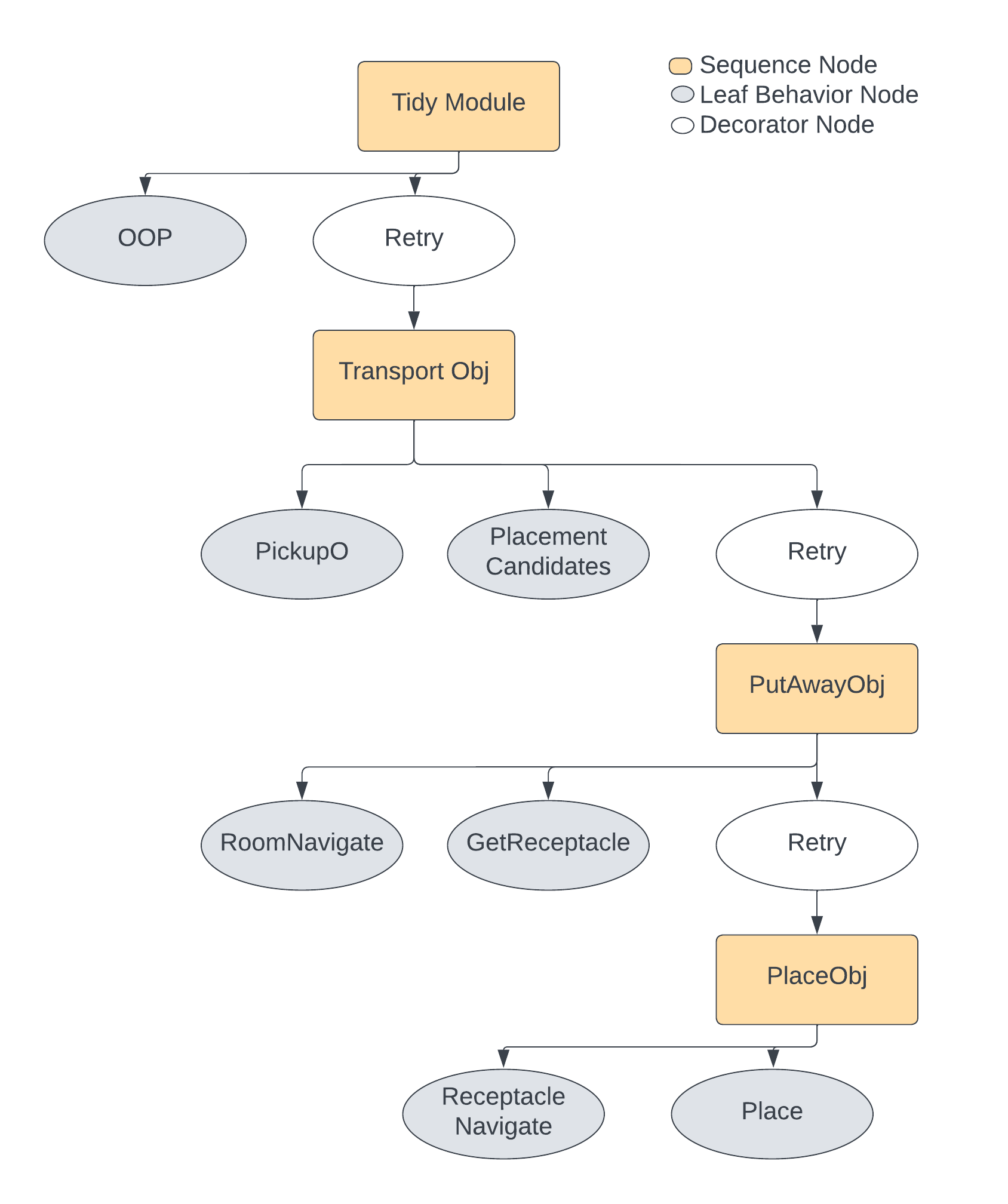}
    \vspace{-2.5mm}
    \caption{The complete behavior tree of the home-robot tidy module}
    \label{fig:tidy_bt}
\vspace{-5pt}
\end{figure} 
Figure \ref{fig:tidy_bt} shows the BT of the home-robot tidy module. The system begins by calling the misplaced object identification (OOP) method of the rearrangement module \ref{sec:oop}. For every object, potential placement candidates (PlacementCandidates) are computed \ref{sec:oop}. The Pickup Behavior \ref{sec:manipulation} is called on the misplaced object. RoomNavigator, followed by ReceptacleNavigator modules are executed, given by the placement candidates. The PlaceBehavior is finally called to place the object. If the place action fails, then the robot tries other candidate receptacles until one succeeds, highlighting the BT's advantages. This is implemented through multiple Decorator Nodes that can facilitate retry behaviors. The different messages from each behavior are passed around through blackboard mechanisms. The visual recognition module constantly runs in the background throughout the episode. The system continues to run until the robot either makes an unrecoverable mistake (such as dropping the object or a hardware failure) or all items are correctly placed. 

% For this work, the $\mathtt{py\_trees}$\footnote{\href{https://github.com/splintered-reality/py\_trees}{https://github.com/splintered-reality/py\_trees}} library is used, due to its pythonic nature, and ROS support. For implementation, as per the behavior tree, sensing and computation nodes are ROS Services while actuation nodes (such as navigate and manipulate) are ROS Actions. The perception node (for objects and receptacle detection) executes continuously in the background and publishes detections at a constant rate, that the different nodes subscribe to.

\begin{figure*}[t]
    \centering
    \includegraphics[width=\textwidth]{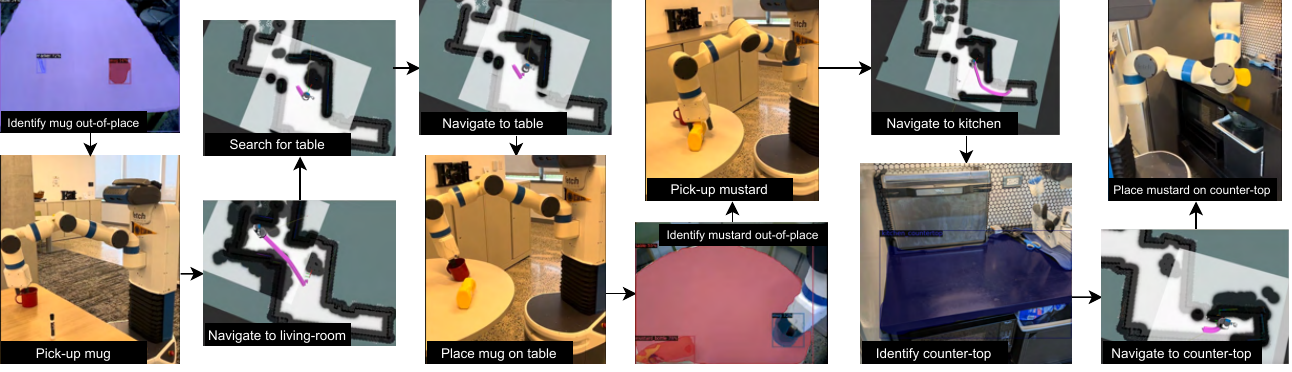}
    \caption{Long horizon rearrangement task. Initially, a $\mathtt{mug}$ is identified to be incorrectly placed on the office table. Then, the robot picks it up, and navigates to the desired target location by first going to the livingroom, and then moving towards the table receptacle. After placing the mug, a second object $\mathtt{mustard\_bottle}$ is found misplaced on the livingroom table. Subsequently, the robot picks the bottle, and transports it to the countertop in the kitchen.}
    \label{fig:long_horizon}
\end{figure*}

\section{ Experiments} \label{sec: expt}

% More object the results of these particual model. Experiments with the rest of the objects. With
% \begin{figure}[t]
%     \centering
%     \includegraphics[width=0.5\textwidth]{images/fetch_all_object_clutter.png}
%     \caption{List of objects considered along with their 2D bounding boxes as given by our detection system}
%     \label{fig:2d_detections}
% \end{figure} 

We test our system through various real-world experiments, involving (i) \textit{Semantic mapping and visual recognition} for generating coarse semantic environment representations and detecting target objects and receptacle surfaces, (ii) \textit{Object rearrangement} for identifying and repositioning misplaced objects, (iii) \textit{Object manipulation} for ensuring stable object interactions, and (iv) \textit{Semantic navigation} for robot's trajectory planning with the generated semantic map.
% We conduct a number of diverse experiments for testing our proposed system in a real-world setting. Each experiment comprises of the four main modules that make up our pipeline -- (i) \textit{Semantic mapping and visual recognition}: For generating coarse semantic representations of the environment, detecting target objects, and receptacle surfaces, (ii) \textit{Object rearrangement}: For determining if any of the detected objects are misplaced, and then predicting their possible target locations, (iii) \textit{Object manipulation}: For facilitating stable grasping of objects for pick-up and placement tasks, and (iv) \textit{Semantic navigation}: For using the generated semantic map to plan the robot's trajectory in the environment. 
Figure \ref{fig:all_comps} contains a pictorial representation of each of our individual modules at work for a tidy-up task. 
% Our mobile manipulation platform consists of a Fetch Mobile Manipulator and Freight Mobile Robot Base by Fetch Robotics \cite{fetch2016}. 
All the experiments are performed in the real world using a simple apartment environment, created from an actual communal office space within our laboratory. The overall environment has an office space, living room, and a kitchen as shown in the semantic map in Figure \ref{fig:all_comps}. The following sections describe the different types of experiments.
% We conduct three fundamental experiments to elucidate the effectiveness of our comprehensive system. The first is what we consider a long-horizon tidy-up task, where the robot must visit and identify at least two misplaced items in two different rooms. The second is a tidy-up task with two trials of the same experimental setup of misplaced items but using two different user preferences for each trial. Lastly, we perform a complex interaction experiment where the robot must return a misplaced item to a closed environment receptacle such as a drawer or fridge. This requires the robot to identify the environmental receptacle, then open and close the receptacle before and after placement. All experiments are performed in the real world using a mock apartment environment, created from an actual communal office space within our Lab. This space consists of a living room, kitchen, and office space and has a total of 14 receptacles. 
% We use the seven object shown in Figure \ref{fig:2d_detections} in our experiments. 

%% TODO (Sanmi) - need to rewrite experiments to decouple setup and what happened

\subsection{Long-horizon object rearrangement} \label{sec: long_hor}
The first experiment that we consider is a long-horizon tidy-up task, where the robot has to identify multiple misplaced objects, and move them to their respective target locations spanning multiple rooms. The rearrangement episode typically begins with detecting a misplaced object, $o_1$, in the environment. The entire tidy module is called to rearrange the object to the correct location. Upon reaching the destination, the robot further scans the environment for any other misplaced objects. If it finds another such object $o_2$, it repeats the entire process sequentially until $o_2$ has also been correctly placed. 
% We illustrate this process in Figure \ref{fig:long_horizon} 
Figure \ref{fig:long_horizon} illustrates the process where $o_1 = \mathtt{mug}$ is transported from an office table to the living-room table, and $o_2 = \mathtt{mustard\_bottle}$ is then moved from the living-room table to the kitchen counter-top.

% The setup for our long horizon experiment is akin to our user preference study, except we select only one user's preference, and all objects are then placed either correctly or incorrectly at random. We use 3 to 5 objects for this experiment and have at least two placed incorrectly. Once the trial begins, the robot must identify objects and their receptacles to determine if the object is misplaced. All misplaced objects must be moved to their correct location based on the user's preference. We leave the robot to explore and move objects until at least two misplaced items in two different rooms are discovered and correctly rearranged. 
\begin{table}[!b]
\scriptsize
\centering
\caption{Preferred object placements for two sampled users}
\label{tb1}
% \begin{tabular}{l|c|c|c|c}\toprule
\begin{tabular}{p{0.8cm}|C{1.1cm}|C{1.8cm}|C{1.1cm}|C{1.8cm}}
\multirow{3}{*}{{\textbf{Objects}}} & \multicolumn{2}{c|}{\textbf{Sampled user $U_1$}} & \multicolumn{2}{c}{\textbf{Sampled user $U_2$}}\\
 & \textbf{Preferred} & \textbf{Preferred} & \textbf{Preferred} & \textbf{Preferred}\\
 & \textbf{rooms} & \textbf{receptacles} & \textbf{rooms} & \textbf{receptacles}\\
\hline
\multirow{3}{*}{\parbox{0.8cm}{rubik's cube}} & office & [shelf, table] &  livingroom & [drawer, table] \\
& kitchen & [counter, table] & office & [table, drawer] \\
& livingroom & [drawer, table] & kitchen & [drawer, table] \\
\hline
\multirow{3}{*}{\parbox{0.8cm}{mustard bottle}} & kitchen & [drawer, counter] & kitchen & [shelf, counter] \\
& livingroom & [table, sofa] & livingroom & [table, drawer] \\
& office & [table, drawer] & office & [drawer, table] \\
\hline
\multirow{3}{*}{\parbox{0.8cm}{marker}} & livingroom & [drawer, shelf] & office & [table, drawer] \\
& office & [table, drawer] & kitchen & [table, drawer] \\
& kitchen & [drawer, table] & livingroom & [table, shelf] \\
\hline
\multirow{3}{*}{\parbox{0.8cm}{cracker box}} & kitchen & [drawer, table] & office & [shelf, drawer] \\
& livingroom & [drawer, table] & kitchen & [drawer, table] \\
& office & [drawer, shelf] & livingroom & [drawer, sofa] \\
\hline
\multirow{3}{*}{\parbox{0.8cm}{bleach cleanser}} & livingroom & [drawer, table] & office & [shelf, table] \\
& office & [shelf, table] & kitchen & [drawer, table] \\
& kitchen & [shelf, drawer] & livingroom & [table, drawer] \\
\hline
\multirow{3}{*}{\parbox{0.8cm}{gelatin box}} & office & [table, shelf] & livingroom & [table, drawer] \\
& kitchen & [drawer, counter] & office & [table, shelf] \\
& livingroom & [drawer, table] & kitchen & [drawer, counter] \\
\hline
\multirow{3}{*}{\parbox{0.8cm}{potted meat can}} & kitchen & [counter, shelf] & office & [drawer, table] \\
& livingroom & [drawer, table] & kitchen & [counter, shelf] \\
& office & [drawer, table] & livingroom & [drawer, table] \\
\hline
\multirow{3}{*}{\parbox{0.8cm}{mug}} & kitchen & [counter, sink] & livingroom & [table, shelf] \\
& livingroom & [shelf, sofa] & office & [drawer, table] \\
& office & [drawer, table] & kitchen & [sink, drawer] \\
% \hline
% \multirow{3}{*}{\parbox{0.8cm}{soup can}} & livingroom & [table, drawer] & office & [drawer, shelf] \\
% & kitchen & [drawer, counter] & kitchen & [drawer, shelf]  \\
% & office & [drawer, table]  & livingroom & [sofa, drawer] \\
\bottomrule
\end{tabular}
\vspace{-10pt}
\end{table}

\subsection{User-preference based object tidy-up}
% Our second experiment considers tidying up an object $o$ in different locations, as the subjective preference of different users. This setup is inspired by the fact that humans typically arrange their homes as-per their choices, and as such, there is seldom a deterministic correct location where objects need to go.
Our second experiment focuses on transferring an object $o$ to different locations, catering to individual user preferences. This experiment acknowledges the subjective nature of object placement in homes. Section \ref{sec:oop} describes a collaborative-filtering approach for generating a user matrix about how objects can be placed differently based on human preference. For this experiment, we sampled two users, $U_1$ and $U_2$, and tabulated their preference regarding target room locations and receptacle surfaces for eight different objects in Table \ref{tb1}.

We conducted real-world experiments using the $\mathtt{mug}$ object. As per Table \ref{tb1}, $U_1$ considers the preferred target room to be $\mathtt{kitchen}$, with the top-2 receptacles surfaces being $\mathtt{counter}$ and $\mathtt{sink}$. In contrast, $U_2$ desires the $\mathtt{mug}$ to be primarily placed in the $\mathtt{livingroom}$, with the top-2 receptacles being $\mathtt{table}$ and $\mathtt{shelf}$. Thus, we perform multiple real-world episodes by sampling the preferences of $U_1$ and $U_2$ as our object rearrangement module, respectively. 
%The rest of the modules corresponding to visual recognition, manipulation, semantic mapping and navigation are similar to that of the long-horizon task in Section \ref{sec: long_hor}.

 % An example of different user preferences is shown in Table \ref{tb1}. For all six of our real-world objects, we show the resulting placements receptacles for two different users based on our module. In this experimental setup up we conduct two trials, selecting users with different object placement preferences for each trial. We use the mug as an example object in our trails.  User A prefers the mug to be placed on the kitchen counter-top or kitchen sink, while User B prefers the mug to be placed on the living room table or office table. According to our model, both users consider their placements as correct placement locations and rank them higher or equal to other possible correct placements. At the start of each experiment trial, the mugs are placed at an incorrect location for both users. The robot must now tidy the environment and place the mug and other objects in the correct location given User A's or User B's preferences. 

\begin{figure}[htb]
    \centering
    \includegraphics[width=0.48\textwidth]{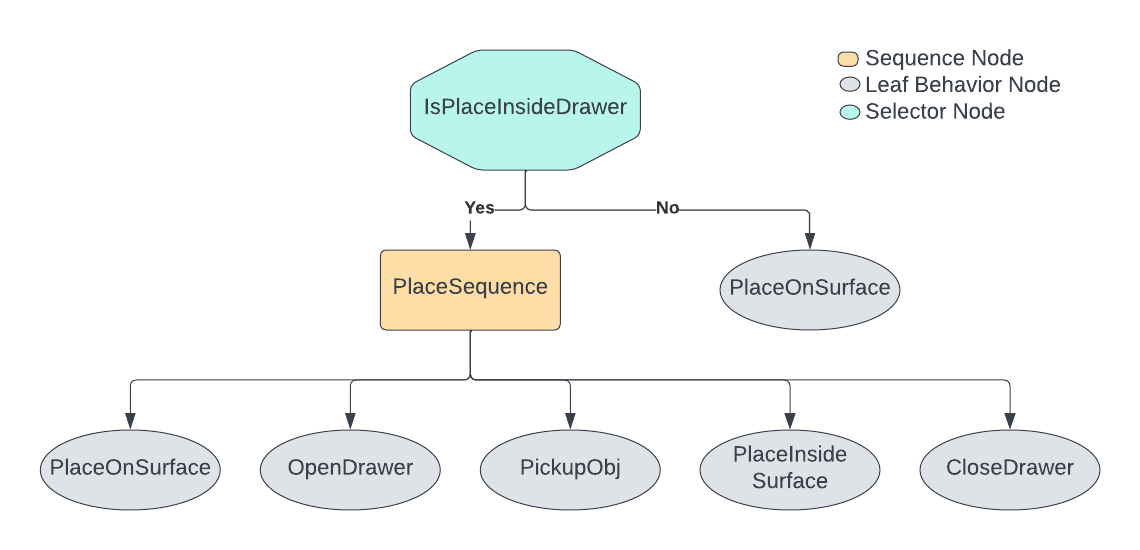}
    \caption{The behavior tree to place an object into the drawer.}
    \label{fig:place_bt}
\end{figure} 

\subsection{Complex interactions} % Jiaming 
\begin{figure}[htb]
    \centering
    \includegraphics[width=0.48\textwidth]{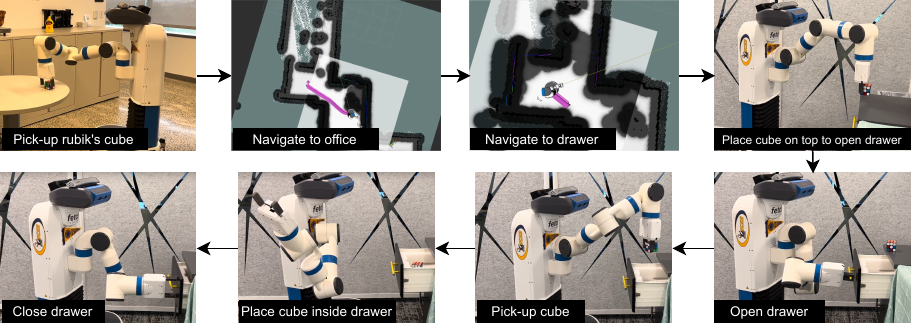}
    \caption{In a multifaceted task such as placing a Rubik's cube into a drawer, a robot must undertake a series of interrelated actions. Initially, the robot approaches the drawer. Recognizing that the drawer must be opened to place the cube inside, it then discerns the need to temporarily set down the Rubik's cube. Only after opening the drawer can it successfully place the cube within.}
    \label{fig:complex_int}
\vspace{-5pt}
\end{figure}

% During a rearrangement task, simply picking and placing objects may not be the only manipulation that a robot requires. Specifically, it is possible that the environment must first be interacted with before enabling the placement task.
A rearrangement task may require the robot to interact with the environment before proceeding with object placement beyond just picking and placing.
For instance, placing an object inside a \textit{closed} receptacle. 
In this work, we demonstrate this concept through the task of placing a Rubik's Cube inside a drawer. Because the drawer is initially closed, the robot has to perform multiple sub-tasks based on the behavior tree shown in Figure \ref{fig:place_bt}. Furthermore, as depicted in Figure ~\ref{fig:complex_int}, the robot estimates a temporary location for the Rubik's Cube and predicts grasp poses to open the drawer. Following that, the robot places the cube into the temporary location and opens the drawer, so it can grasp and place the cube inside the drawer. 
% This example highlights the importance of flexibility and adaptability in robotic manipulation, ensuring the robot can interact successfully with a complex environment. 
%todo (siyer) - add a BT for this maybe?

% More object the results of these particular model. Experiments with the rest of the objects. With

\section{Summary} \label{sec: conclusion}

The world needs a home robot that can do more than vacuuming. We have presented key components for navigating robustly in a home setting, for detecting of objects and receptacles and determining if they are out of place. Skills for manipulating and handling objects in a daily setting for a task such as clean-up or reset of a home to a nominal setting are introduced to allow clean-up. Finally, using a combination of common-sense reasoning and recommender systems, a strategy to detect objects out of place and suggest improved locations to put them is discussed. All these techniques are integrated into a consistent and robust framework using behavior trees and implemented on the Fetch robot using a ROS based architecture. We have demonstrated the final system and how it can be used in a real-world scenario with modest complexity, for clean-up of a space by placement of objects in appropriate locations. 

% Clearly, there is a need for longer-term testing across multiple environments, so this is mainly a demonstration of the integration of a broad set of basic skills to provide end-to-end functionality. In addition, it is of interest to optimize the system to make it fast enough to compete with humans in clean-up of a space. 
In this work, we mainly demonstrated an integration of fundamental robotic skills in a modular representation for house-hold tidy-up tasks. In the future, we want to aim for longer-term testing across multiple environments, with an additional goal of optimizing the system for speed in space cleanup, such that it can compete with humans.

\bibliographystyle{IEEEtran}
\bibliography{bibliography}

\end{document}